%% file: template.tex
\documentclass{article}
\usepackage{arxiv}

\usepackage{soul}
\usepackage[utf8]{inputenc} 
\usepackage[T1]{fontenc}    
\usepackage{hyperref}       
\usepackage{url}            
\usepackage{booktabs}       
\usepackage{amsfonts}       
\usepackage{nicefrac}       
\usepackage{microtype}      
\usepackage{lipsum}		
\usepackage{graphicx}
\usepackage{array}
\usepackage{doi}
\usepackage{multirow}
\usepackage{multicol}
\usepackage{makecell}
\usepackage{xcolor}
\usepackage{subcaption}
\usepackage[normalem]{ulem}
\usepackage{amsmath,amsfonts,bm}
\usepackage{amssymb,amsthm}
\usepackage{enumitem}
\usepackage{caption}
\usepackage{subcaption}
\usepackage{adjustbox}
\usepackage{afterpage}
\usepackage[most]{tcolorbox}
\usepackage{listings}
\usepackage{wrapfig} 

\newtcblisting{mybox}[1]{
    colback=gray!5,           
    colframe=black!75,        
    fonttitle=\bfseries\sffamily, 
    title=#1,                 
    enhanced,                 
    attach boxed title to top left={yshift=-2mm, xshift=2mm}, 
    boxed title style={sharp corners=south, colback=black!75}, 
    sharp corners=false,      
    arc=3pt,                  
    left=10pt, right=10pt, top=10pt, bottom=10pt, 
    listing only,             
    listing options={
        basicstyle=\ttfamily\small,
        breaklines=true,
        columns=fixed,
        keepspaces=true,
        extendedchars=false
    }
}

\usepackage[
    backend=biber,
    citestyle=authoryear,
    bibstyle=authortitle,
    mincitenames=1,
    maxcitenames=1,
    maxbibnames=3
]{biblatex}
\definecolor{hy}{RGB}{255, 128, 0}

\newcommand{\citep}[1]{\parencite{#1}}
\renewcommand{\thefootnote}{\fnsymbol{footnote}}
\setlist[itemize,1]{leftmargin=18pt, noitemsep, topsep=3pt}
\setlist[enumerate,1]{leftmargin=18pt, noitemsep, topsep=3pt}
\newcommand{\ourname}{{WorldVQA}}
\newcommand{\ourmodelname}{{Kimi K2.5}}
\renewcommand{\undertitle}{}
\renewcommand{\shorttitle}{\raisebox{-0.12\height}{\includegraphics[width=0.02\textwidth]{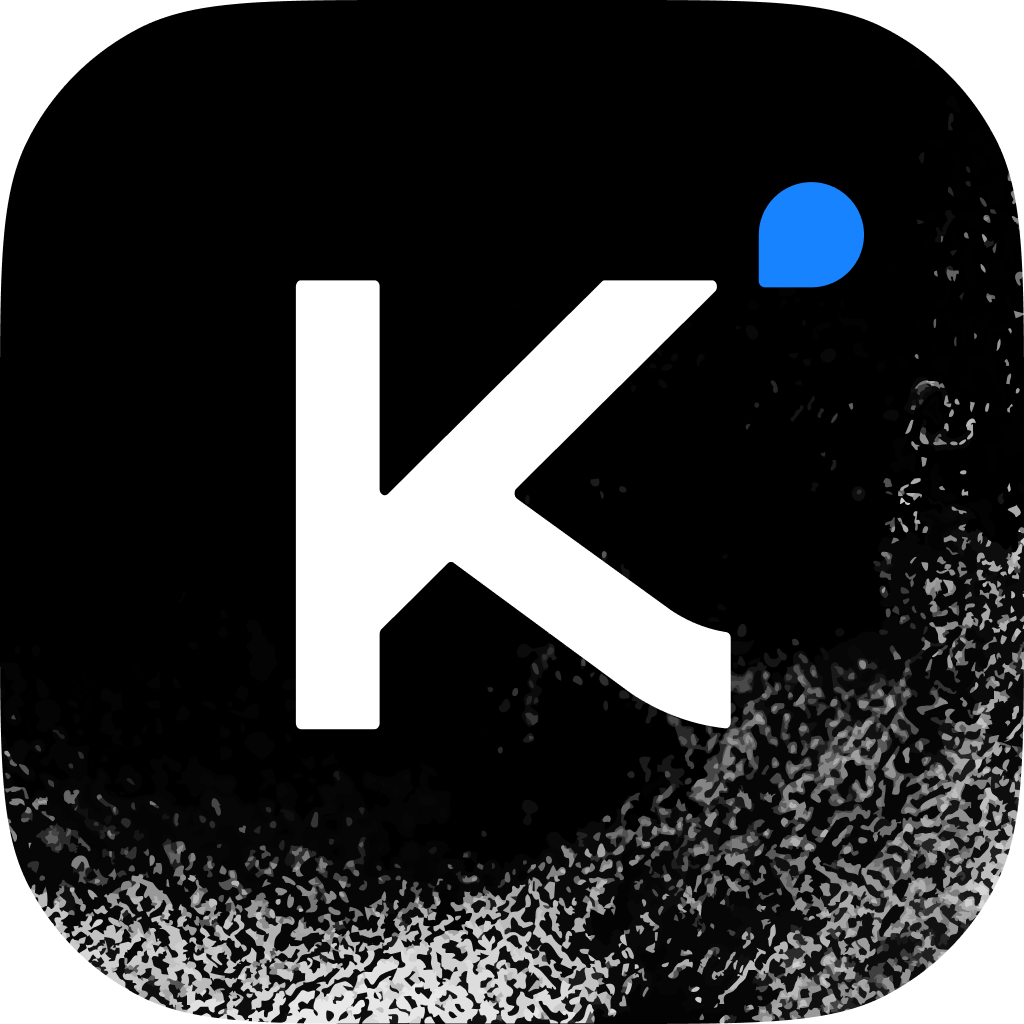}}}
\title{\ourname: Measuring Atomic World Knowledge in Multimodal Large Language Models}

\author{
\\[-5mm]
\textbf{Runjie Zhou}$^{1*}$ \quad
\textbf{Youbo Shao}$^{1*}$ \quad
\textbf{Haoyu Lu}$^{1*\ddag}$ \quad
\textbf{Bowei Xing}$^{1}$ \quad
\textbf{Tongtong Bai}$^{1}$ \quad
\textbf{Yujie Chen}$^{1}$ \quad  \\
\textbf{Jie Zhao}$^{1}$ \quad 
\textbf{Lin Sui}$^{1}$ \quad 
\textbf{Haotian Yao}$^{1}$ \quad 
\textbf{Zijia Zhao}$^{1}$ \quad 
\textbf{Hao Yang}$^{1}$ \quad 
\textbf{Haoning Wu}$^{1}$ \quad
\textbf{Zaida Zhou}$^{1}$ \quad \\
\textbf{Jinguo Zhu}$^{1}$ \quad
\textbf{Zhiqi Huang}$^{1}$ \quad
\textbf{Yiping Bao}$^{1}$ \quad
\textbf{Yangyang Liu}$^{1}$ \quad
\textbf{Y.Charles}$^{1}$ \quad 
\textbf{Xinyu Zhou}$^{1}$ \quad
\\[2ex]
$^1$ Moonshot AI
}

\date{}

\begin{document}

\maketitle
\let\thefootnote\relax\footnotetext{$^*$ Equal contribution.  $^\ddag$ Project lead.}

\vspace{-10pt}
\input{sec/0_abstract}
\input{sec/1_intro}
\input{sec/2_WorldVQA}
\input{sec/3_Experiments}
\input{sec/4_Related_Work_and_Discussion}
\printbibliography[title={References}]
\appendix
\input{sec/Appendix}
\end{document}

%% file: sec/0_abstract.tex
\begin{abstract}
We introduce \ourname, a benchmark designed to evaluate the atomic visual world knowledge of Multimodal Large Language Models (MLLMs).
Unlike current evaluations, which often conflate visual knowledge retrieval with reasoning, \ourname~decouples these capabilities to strictly measure "what the model memorizes."
The benchmark assesses the atomic capability of grounding and naming visual entities across a stratified taxonomy, spanning from common head-class objects to long-tail rarities.
We expect \ourname~to serve as a rigorous test for visual factuality, thereby establishing a standard for assessing the encyclopedic breadth and hallucination rates of current and next-generation frontier models. The dataset can be found in \href{https://worldvqa2026.github.io/WorldVQA/#}{WorldVQA Homepage}.
\end{abstract}

%% file: sec/1_intro.tex
\section{Introduction}
\label{sec:intro}

\begin{figure}[h]
    \centering
    \includegraphics[width=.9\linewidth]{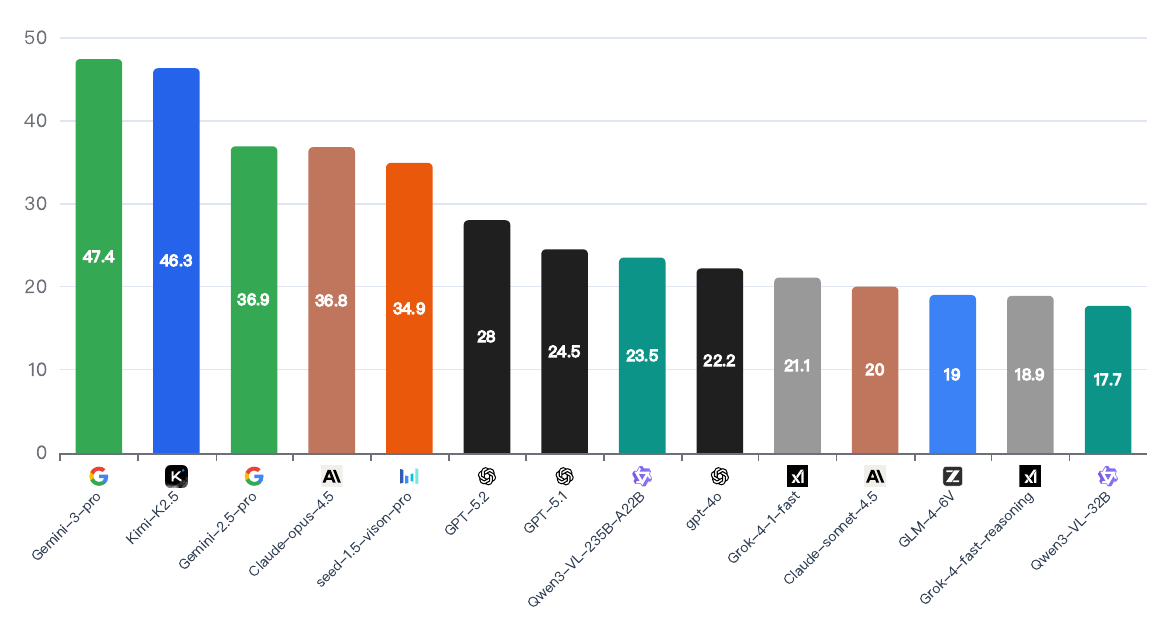}
    \caption{\textbf{Overall Model Accuracy on \ourname}. While the Gemini-3-pro (47.4\%) and \ourmodelname~(46.3\%) currently lead the field, no evaluated model surpasses the 50\% accuracy threshold, underscoring the significant challenge of grounding atomic visual knowledge.}
    \label{fig:WorldVQA_Charts}
\end{figure}

\begin{figure}[h]
    \centering
    \includegraphics[width=\linewidth]{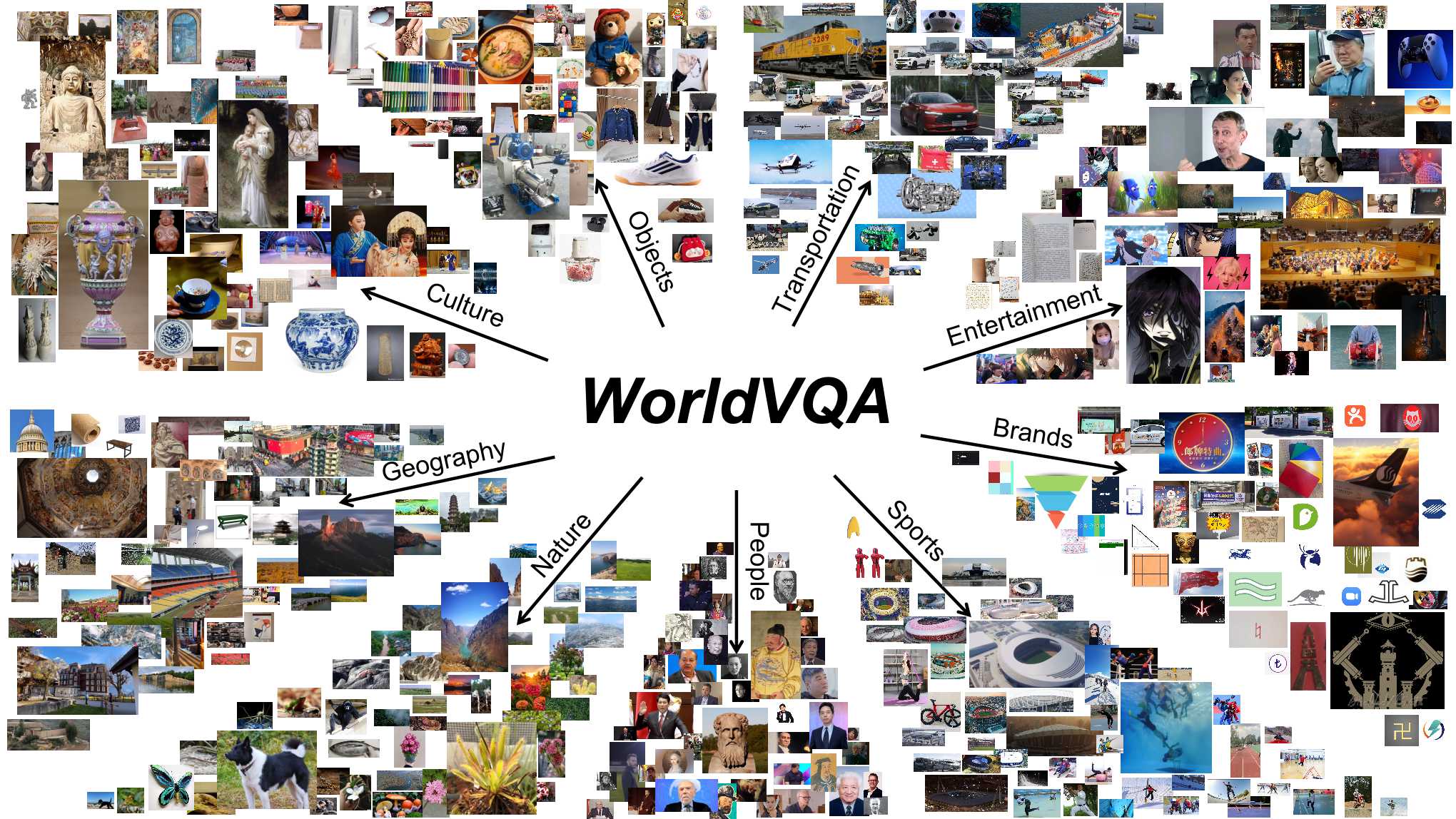}
    \caption{A visual overview of the \textbf{WorldVQA} dataset. The benchmark is organized into nine categories: Nature \& Environment (Nature); Locations \& Architecture (Geography); Culture, Arts \& Crafts (Culture); Objects \& Products (Objects); Vehicles, Craft \& Transportation (Transportation); Entertainment, Media \& Gaming (Entertainment); Brands, Logos \& Graphic Design (Brands); Sports, Gear \& Venues (Sports); Notable People \& Public Figures (People). The visual entities curated to evaluate atomic world knowledge range from globally recognized "head-class" landmarks and logos to specific "long-tail" biological species and artisanal artifacts. To maintain  \textbf{atomic isolation}, each image serves as an unambiguous visual stimulus for entity naming, strictly decoupled from complex reasoning or OCR dependencies.}
    \label{fig:main_figure}
\end{figure}

The advancement of MLLMs is contingent upon two distinct capabilities: reasoning (processing logic and relationships) and knowledge grounding (mapping sensory inputs to factual reality). While recent frontier models have demonstrated encyclopedic mastery in the textual domain, their ability to maintain factual reliability in the visual domain remains inconsistent. True multimodal intelligence requires a robust visual encyclopedia, an internal representation that accurately maps raw pixel data to specific entity identities. Without this precise image grounding, MLLMs function as descriptive engines rather than knowledgeable observers, resulting in a propensity for visual hallucinations where models fabricate plausible but incorrect details.

Accurately quantifying a model's internal visual knowledge is difficult because current evaluation methods fail to isolate visual recognition from high-level cognitive reasoning. Existing Visual-Question-Answering (VQA) benchmarks, such as MMMU (\cite{yue2024mmmu}) and MMStar (\cite{chen2024we}), prioritize complex, multi-step reasoning. These benchmarks conflate visual knowledge with logical deduction, making it difficult to isolate the source of error. Meanwhile, VQA knowledge benchmarks such as SimpleVQA (\cite{cheng2025simplevqa}) often couple visual recognition with secondary dependencies, such as language knowledge or optical character recognition (OCR). For instance, a failure to answer a question about a company's founding date may stem from a lack of textual historical knowledge rather than a failure to visually identify the company's logo. This entanglement prevents researchers from determining whether a deficit lies in the model's visual perception (the "eyes") or its semantic memory (the "brain").


To address this gap, we introduce \ourname, a targeted evaluation suite comprising 3,500 VQA pairs across 9 semantic categories (see Figure \ref{fig:main_figure}). Unlike existing visual knowledge benchmarks that couple visual knowledge and reasoning, \ourname~is engineered to assess \textbf{atomic world knowledge}, the direct, unassisted association between a visual stimulus and its specific proper noun or taxonomic name. To provide a clearer contrast, Table \ref{tab:benchmark_comparison} summarizes the key differences between WorldVQA and these representative benchmarks in multiple dimensions, highlighting our focus on isolating atomic vision concepts.

\ourname~follows four technical design principles:

\textbf{1. Atomic Isolation.} To eliminate confounding variables, we strictly isolate the target task of entity naming. The benchmark evaluates the most fundamental unit of visual knowledge by decoupling visual identification from external logic. We systematically exclude queries requiring OCR, arithmetic, or multi-hop knowledge retrieval. By focusing exclusively on visual grounding (e.g., \textit{“What is the specific scientific name of this species?”}), we isolate the model’s visual recognition capabilities from its reasoning engine.

\textbf{2. Taxonomic Diversity.} To rigorously test knowledge breadth, we synthesized a classification framework covering 9 primary categories (as visualized in Figure \ref{fig:main_figure}): \textit{Nature \& Environment (Nature); Locations \& Architecture (Geography); Culture, Arts \& Crafts (Culture); Objects \& Products (Objects); Vehicles, Craft \& Transportation (Transportation); Entertainment, Media \& Gaming (Entertainment); Brands, Logos \& Graphic Design (Brands); Sports, Gear \& Venues (Sports); Notable People \& Public Figures (People)}. This taxonomy moves beyond unstructured web-crawling, ensuring a balanced distribution that tests both high-frequency "head" entities and the "long-tail" of rare instances. This structure allows us to profile the "encyclopedic boundary" of a model's visual memory.

\textbf{3. Data Integrity \& Verification.} We implement a multi-stage curation pipeline to ensure \ourname~serves as a gold standard oracle. To prevent contamination, we perform rigorous deduplication against common large-scale pre-training corpora. Verification utilizes a dual-gate mechanism combining automated consistency checks via high-performance MLLMs with human-in-the-loop validation, minimizing label noise and ensuring high-fidelity evaluation.

\textbf{4. High Performance Headroom.} The benchmark is calibrated to challenge current frontier models. As illustrated in Figure \ref{fig:WorldVQA_Charts}, even advanced systems exhibit performance ceilings, often failing to exceed 50\% accuracy across all categories. The radar charts (see Figure \ref{fig:radar_comparison}) reveal distinct knowledge pits, particularly in \textit{Nature} and \textit{Culture}, where model performance lags significantly behind text-only equivalents. This demonstrates that \ourname~provides high-resolution visibility into the limitations of current visual pre-training.

By isolating atomic knowledge retrieval from reasoning, \ourname~provides a precise metric for visual hallucination and knowledge grounding. We release this benchmark to the community as a standard for assessing the factual reliability of next-generation MLLMs.

\begin{table}[htbp]
\centering
\resizebox{\textwidth}{!}{
\begin{tabular}{cccc>{\raggedright\arraybackslash}p{3.6cm}p{8cm}}
\toprule
\textbf{Benchmark} & \textbf{Data Size} & \textbf{Language} & \textbf{Question Type} & \multicolumn{1}{c}{\textbf{Target Domain}} & \multicolumn{1}{c}{\textbf{Detail}} \\ \midrule
MMMU-Pro & 5.2k & EN & Multiple-Choice & Academic Understanding & Evaluates expert-level academic knowledge, often conflating factual recall with complex logical reasoning. \\
MMBench & 2.4k & CN \& EN & Multiple-Choice & General Multi-modal Ability & Assesses various multimodal abilities by using perception and reasoning as its primary evaluation pillars. \\
RealWorldQA & 765 & EN & Multiple-Choice & Spatial \& Physical Perception & Measures understanding of physical environments and spatial relationships through situational queries. \\
SimpleVQA & 2.0k & CN \& EN & Generation & Vision Knowledge \& Reasoning & Probes the factuality ability of MLLMs to answer natural language short questions. \\ \midrule
\textbf{WorldVQA} & 3.5k & CN \& EN & Generation & Atomic Vision Knowledge & Isolates atomic world knowledge with stratified, encyclopedic taxonomy. \\ 
\bottomrule
\end{tabular}
}
\caption{Comparison of \ourname~with existing Multimodal Benchmarks. While existing suites conflate factual recall with reasoning or secondary dependencies , WorldVQA stands apart by strictly isolating atomic visual knowledge through the principle of decoupling.}
\label{tab:benchmark_comparison}
\end{table}

%% file: sec/2_WorldVQA.tex
\section{Data Collection and Verification}
The construction of \ourname~follows a rigorous pipeline designed to ensure atomic factuality and taxonomic breadth. The process comprises two primary phases. First, 10 expert annotators with over one year of experience in MLLMs evaluation curated the atomic entities and VQA pairs according to strict taxonomic and granularity standards. Second, these samples underwent a dual-verification process, including automated fact-checking by frontier MLLMs to assess visual clarity and independent blind validation by expert annotators to ensure absolute ground-truth reliability. Only samples passing all verification stages were included in the final benchmark.
\subsection{Design Principles and Criteria}
\ourname~is strictly anchored by the principle of Atomic Isolation. Unlike benchmarks that conflate visual recognition with multi-hop reasoning, \ourname~isolates the model's ability to ground specific visual entities. Beyond this core tenet, we prioritize Encyclopedic Diversity and Data Fidelity to guarantee data reliability. To achieve these, we enforce four strict criteria:

\textbf{Atomic Isolation} To evaluate world knowledge ability, we strictly decouple knowledge retrieval from multi-step reasoning. Questions are engineered to be single-hop, requiring only the direct identification of a visual entity or its specific attributes. Tasks involving OCR, arithmetic, or external logical deduction are excluded to ensure the evaluation reflects the model's internal parametric memory.

\textbf{Encyclopedic Knowledge Coverage} To ensure the benchmark serves as a global standard, data composition is governed by rigorous distribution rules rather than random sampling. First, each category maintains a sufficient volume of data to ensure statistical significance. Second, we prioritize cultural diversity by capping region-specific entities; specifically, the proportion of entities unique to the Chinese context is limited to under 50\% for each individual category, resulting in a final aggregate of 36\% Chinese-specific entities across the entire benchmark. Third, rather than selecting entities at random, annotators were instructed to deliberately sample entities across a broad spectrum of real-world prevalence. These distribution rules ensure that \ourname captures the natural distribution of encyclopedic knowledge, spanning from globally ubiquitous head-class concepts to highly specialized long-tail rarities, as empirically validated by our difficulty alignment analysis in Section \ref{sec: difficulty alignment}.

\textbf{Granularity Alignment} To guarantee the validity of knowledge, we enforce a strict alignment between the specificity of the question and the granularity of the answer. Correctness is defined by taxonomic precision: for example, if an image depicts a Bichon Frise, the answer must identify the specific breed, whereas generic hypernyms such as dog are considered incorrect. This constraint serves two critical functions: it prevents models from bypassing the knowledge requirement through safe but vague generalizations, and it aligns the difficulty level with the true complexity of the visual signal.

\textbf{Visual Reliability} Images must serve as authentic, definitive evidence for the atomic fact being tested. We strictly enforce two standards for visual selection:

    \begin{itemize}
        \item \textbf{Sanitization}: Images must be devoid of textual leakage (e.g., labels, watermarks, overlay text) to preclude the model from using OCR-based shortcuts to "read" the answer.
        \item \textbf{Unambiguity}: The visual features must be distinct and strictly correspond to the target entity. The image must firmly support the ground truth while ruling out reasonable alternatives or confusing distractors. If an entity cannot be uniquely identified from the visual features alone, it is discarded.
    \end{itemize}

\subsection{Data Quality}
The integrity of a benchmark is determined by the precision of its data. To ensure \ourname~serves as a definitive measure of atomic world knowledge, we implemented a rigorous curation and verification pipeline.

\subsubsection{Data Curation Pipeline}
We employed a three-step pipeline to collect raw taxonomic entities and convert entities into high-quality VQA triplets:

\textbf{Step 1: Seed Entity Collection.} 10 annotators gathered initial data based on the taxonomy. Seed entities were sourced from internal lexicons. Adhering to the \textit{Encyclopedic Knowledge Coverage} criteria, annotators selected appropriate entities and retrieved corresponding images from trusted web sources (following the \textit{Visual Reliability} criteria), finally formulating QAs according to the \textit{Granularity Alignment} criteria.

\textbf{Step 2: Distributional Balancing and Global Expansion.} To ensure the benchmark's international generalizability, we perform contextual labeling to identify and partition region-specific entities. We enforce a 50\% per-category cap on entities unique to the Chinese context. For categories falling below our global representation threshold, we utilize an LLM-in-the-loop expansion strategy, leveraging GPT and Kimi for association search, to identify supplemental global entities, which are then integrated following the protocol in Step 1.

\textbf{Step 3: Visual Deduplication.} To eliminate redundancy and mitigate data leakage from common pre-training corpora, we employed the Instance-level Semantic Content (ISC, \cite{yokoo2021contrastivelearninglargememory}) descriptor to detect near-duplicate images. For each candidate image, we calculated the cosine similarity of ISC embeddings for each candidate image against massive open-source datasets, specifically LAION (\cite{schuhmann2022laion5bopenlargescaledataset}) and Common Crawl. Applying a strict threshold of 0.95, any images identified as duplicates or leaked from these sources were discarded. To maintain data volume without duplication, we performed targeted re-collection for these entities by capturing new visual assets from video screenshots, which minimizes the likelihood of the model relying on memorized image-answer pairs. This rigorous protocol ensures that correct responses reflect the model’s genuine internal encyclopedia rather than simple pattern retrieval from its training history.

\begin{table}[!htbp]
\centering
\small
\setlength{\tabcolsep}{8pt} 
\begin{tabular}{lr|lr}
\toprule
\multicolumn{2}{c|}{\textbf{Dataset Composition}} & \multicolumn{2}{c}{\textbf{Semantic Taxonomy}} \\
\cmidrule(r){1-2} \cmidrule(l){3-4}
\textbf{Metric} & \textbf{Value} & \textbf{Category} & \textbf{Ratio} \\ 
\midrule
\textbf{Total Samples} & \textbf{3500} & Locations \& Architecture & 14.63\% \\
 & & Entertainment, Media \& Gaming & 14.60\% \\
\multicolumn{2}{l|}{\textbf{Language}} & Culture, Arts \& Crafts & 14.46\% \\
- English (EN) & 64.00\% & Notable People \& Public Figures & 14.29\% \\
- Chinese (CN) & 36.00\% & Objects \& Products & 12.49\% \\
\multicolumn{2}{l|}{\textbf{Difficulty}} & Nature \& Environment & 9.31\% \\
- Easy & 31.16\% & Vehicles \& Transportation & 8.74\% \\
- Medium & 40.77\% & Brands, Logos \& Design & 7.43\% \\
- Hard & 28.07\% & Sports, Gear \& Venues & 4.06\% \\
\bottomrule
\end{tabular}
\caption{\ourname~statistics. \textbf{Left:} Core dataset composition including language split and difficulty stratification. \textbf{Right:} Distribution across the nine semantic categories, sorted by prevalence.}
\label{tab:statistics}
\end{table}

\subsection{Model-Based Difficulty Stratification}
To ensure high discriminative capacity and mitigate the ceiling effect observed in existing benchmarks, we applied a Model-Performance-Based Stratification strategy.

We evaluated all candidate samples using an ensemble of five frontier MLLMs. To maximize the benchmark's utility and discriminative power, we discarded trivial samples correctly answered by all five models. The remaining samples were stratified into three difficulty levels based on model performance: Easy (>3 models correct), Medium (1–2 models correct), and Hard (0 models correct). To prevent the benchmark from being dominated by simpler entities and to maintain a focus on challenging long-tail knowledge, we performed random downsampling on the Easy category. We report the final proportions of each difficulty tier in Table \ref{tab:statistics}.

\textbf{Note on Bias.} This stratification is not intended to "trap" specific models, but rather to counteract the ceiling effect prevalent in current benchmarks. By intentionally downsampling from the model-defined Easy tier, we ensure \ourname~remains a challenging probe for next-generation frontier systems. Meanwhile, all Hard samples underwent a mandatory secondary human review to confirm that the difficulty stems from the rarity of the knowledge, not from visual ambiguity or annotation error.

\subsection{Dual-Verification Mechanism}
To ensure maximum ground-truth fidelity, we implement a rigorous dual-gate verification protocol consisting of automated model-based auditing and independent human validation. This multi-stage process is designed to filter out semantic noise and visual ambiguity.

\textbf{Model Based Visual Auditing.} We utilize few-shot prompted Gemini-3-Pro as automated fact-checker to evaluate each VQA triplet (see Visual Audit Prompt in  Appendix \ref{sec:prompt}). The models enforce three non-negotiable requirements for data integrity: 
\begin{itemize} 
\item \textbf{Visual Clarity}: The image must provide sufficient resolution to permit unambiguous entity identification. 
\item \textbf{Semantic Exclusivity}: The image content must uniquely support the ground-truth label while actively ruling out reasonable alternative interpretations or distractors. 
\item \textbf{Contextual Completeness}: The visual context must encompass all necessary information required to resolve the question. 
\end{itemize}

\textbf{Human Blind Validation.} Parallel to automated checks, we conducted independent human validation. An annotator, unaware of the ground truth, was required to answer the question. Any sample where the human prediction diverged from the ground truth was flagged for manual audit. Cases involving factual errors or visual ambiguity were permanently purged from the dataset.

\subsection{Grading and Metrics}
To facilitate a standardized comparison, we utilize Accuracy as our primary single-number metric to measure overall factual reliability. For a more granular experimental analysis, we also report Correct Given Attempted (CGA) and F-score. CGA isolates the precision of the model’s internal knowledge by evaluating only the samples it chose to answer, effectively measuring its susceptibility to hallucination when it commits to a response. The F-score synthesizes coverage (attempt rate) and precision (CGA) into a single harmonic mean, penalizing both over-conservative refusal and over-aggressive guessing.

\subsection{Benchmark Statistics}
\label{sec:stats}

Following the curation protocols described above, we present a high-level overview of the \ourname~dataset. As shown in Table \ref{tab:statistics}, the benchmark consists of 3,500 pairs with a balanced linguistic and categorical spread. Importantly, to establish a global evaluation standard, we maintain a 1:1.78 Chinese-to-English ratio.

%% file: sec/3_Experiments.tex
\section{Experiments}
\subsection{Settings}
To ensure a rigorous and fair evaluation across the diverse landscape of MLLMs, we maintained strict consistency in our experimental protocols. All models evaluated with unified prompts and official inference parameters. For the grading process, we employed GPT-oss-120b (\cite{openai2025gptoss120bgptoss20bmodel}) as our primary judge model (see Appendix \ref{sec:prompt} for the judge prompt). To validate this automated grading, a manual audit of 160 random samples reveals a 98.1\% alignment rate with human expertise (only 3 disagreements).

\subsection{Main Results}
\begin{table}[htbp]
\centering
\small
\setlength{\tabcolsep}{3pt}
\resizebox{\textwidth}{!}{ 
\begin{tabular}{l|cccc|ccccccccc}
\toprule
\multirow{2}{*}[-3ex]{\textbf{Models}} & \multicolumn{4}{c|}{\textbf{Overall results}} & \multicolumn{9}{c}{\textbf{F-score on 9 task categories}} \\
\cmidrule(lr){2-5} \cmidrule(lr){6-14}
 & Accuracy & \makecell{Not \\ Attempted} & \makecell{Correct \\ Given \\ Attempted} & F-score & Nature & \makecell{Geo- \\ graphy} & Culture & Objects & \makecell{Trans- \\ portation} & \makecell{Entertain- \\ ment} & Brands & Sports & People \\
\midrule
\multicolumn{14}{c}{\textbf{Closed-source MLLMs}} \\
\midrule
Gemini-3-pro & 47.4 & 0.6 & 47.7 & 47.5 & 45.1 & 44.7 & 47.2 & 48.1 & 45.1 & 47.6 & 52.4 & 59.4 & - \\
Gemini-2.5-pro & 36.9 & 0.1 & 36.9 & 36.9 & 37.1 & 33.8 & 32.6 & 39.6 & 39.9 & 34.2 & 38.8 & 54.2 & - \\

Seed-1.5-vision-pro & 34.9 & 1.6 & 35.5 & 35.2 & 41.4 & 36.1 & 33.4 & 32.8 & 35.0 & 33.6 & 32.3 & 43.7 & - \\

Claude-opus-4.5 & 36.8 & 3.4 & 38.1 & 37.5 & 32.5 & 36.5 & 34.1 & 39.6 & 43.5 & 29.0 & 47.6 & 54.9 & - \\
Claude-sonnet-4.5 & 20.0 & 8.0 & 21.8 & 20.9 & 19.4 & 21.0 & 17.4 & 22.9 & 24.8 & 11.6 & 32.2 & 31.0 & - \\

GPT-5.2 & 28.0 & 5.4 & 29.5 & 28.7 & 24.3 & 29.1 & 26.7 & 26.6 & 30.7 & 24.8 & 39.1 & 40.8 & - \\
GPT-5.1 & 24.5 & 16.3 & 29.3 & 26.7 & 27.3 & 25.1 & 22.5 & 26.6 & 31.6 & 18.5 & 36.0 & 45.4 & - \\
GPT-4o & 22.2 & 9.1 & 24.4 & 23.3 & 25.6 & 20.6 & 17.8 & 19.1 & 26.2 & 19.1 & 35.2 & 44.5 & - \\

Grok-4-1-fast-reasoning & 21.1 & 0.1 & 21.1 & 21.1 & 18.4 & 23.6 & 20.2 & 25.2 & 23.5 & 11.4 & 25.8 & 30.3 & - \\
Grok-4-fast-reasoning & 18.9 & 0.2 & 19.0 & 18.9 & 17.8 & 19.0 & 18.6 & 22.0 & 20.3 & 8.3 & 26.6 & 34.5 & - \\

\midrule
\multicolumn{14}{c}{\textbf{Open-source MLLMs}} \\
\midrule
\ourmodelname & 46.3 & 2.1 & 47.3 & 46.8 & 40.6 & 46.8 & 43.0 & 44.7 & 47.4 & 48.1 & 52.6 & 64.8 & 50.9 \\
Kimi-VL-16B-A3B & 12.0 & 3.3 & 12.4 & 12.2 & 11.2 & 13.9 & 10.1 & 10.8 & 13.5 & 7.9 & 20.8 & 17.7 & 7.4 \\
Qwen3-VL-235B-A22B-Instruct & 23.5 & 0.0 & 23.5 & 23.5 & 26.1 & 24.8 & 22.9 & 26.1 & 28.8 & 15.5 & 22.3 & 26.1 & 26.2 \\
Qwen3-VL-32B-Instruct & 17.7 & 0.0 & 17.7 & 17.7 & 18.1 & 18.0 & 16.8 & 19.0 & 19.0 & 12.1 & 23.8 & 20.4 & 13.1 \\
GLM-4.6V & 19.0 & 0.0 & 19.0 & 19.0 & 24.5 & 21.5 & 17.8 & 19.2 & 18.6 & 12.5 & 20.4 & 23.2 & 10.7 \\
GLM-4.6V-Flash & 14.8 & 0.1 & 14.8 & 14.8 & 16.0 & 16.3 & 13.2 & 14.9 & 19.0 & 7.8 & 18.8 & 20.4 & 8.2 \\

\bottomrule
\end{tabular}}
\caption{Performance of frontier MLLMs on \ourname. Hyphen entries (-) denote scores omitted due to excessive refusal rates. Overall Results aggregate the first eight categories. "Notable People \& Public Figures" (People) is excluded from the overall average to ensure a fair comparison, as systematic refusals in closed-source models, driven by privacy and safety guardrails, do not necessarily reflect underlying knowledge deficits.}
\label{main result}
\end{table}

\begin{figure}[htbp]
  \centering
  \includegraphics[width=.65\linewidth]{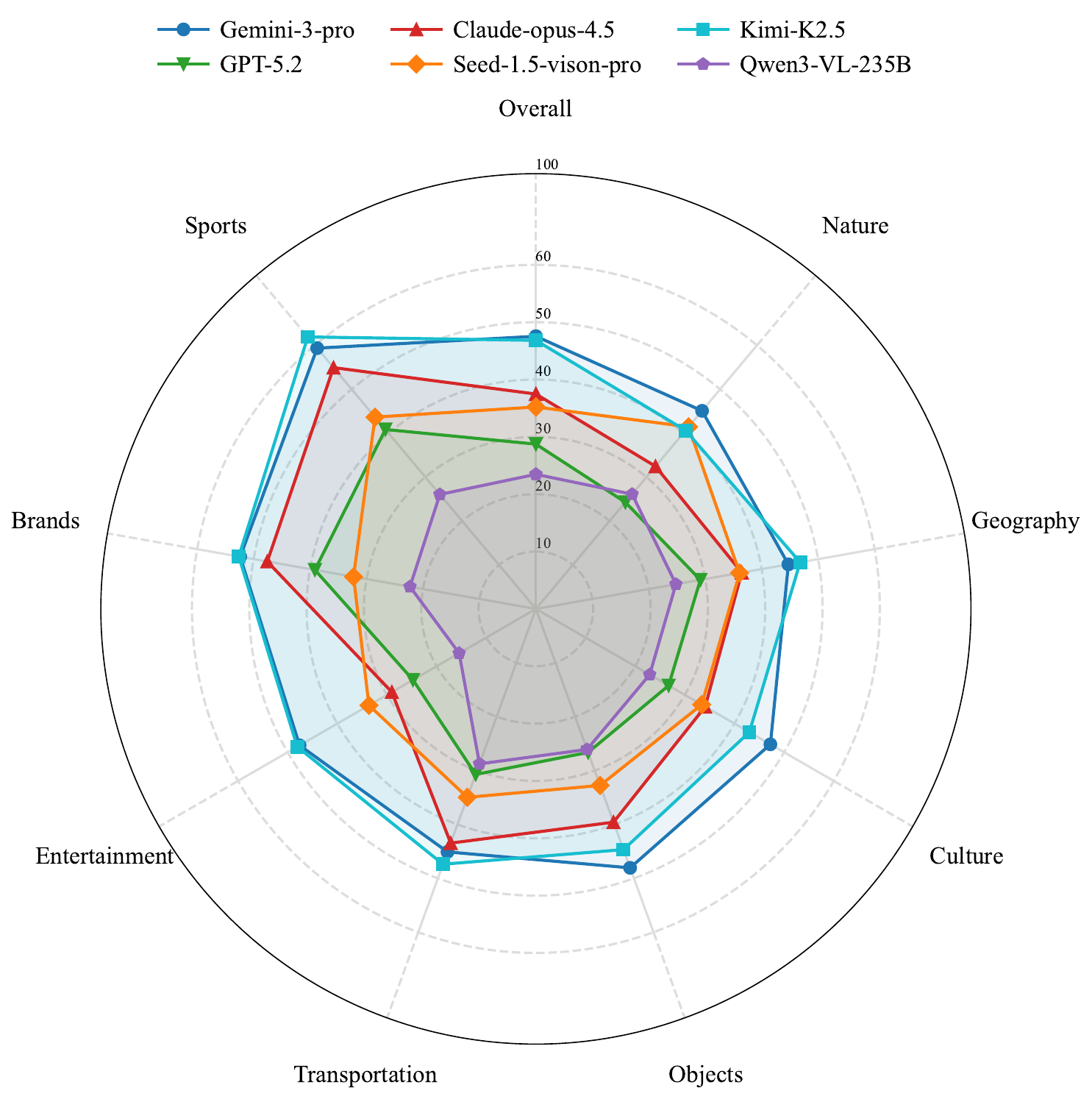}
  \caption{\textbf{Category-wise F-score comparison on WorldVQA.} This radar chart illustrates the performance profiles of frontier close-source and open-source MLLMs across the 8 semantic categories. The visualization highlights the relative proficiency in high-frequency domains like \textit{Sports} and \textit{Brands}, while revealing significant performance troughs in specialized domains such as \textit{Nature} and \textit{Culture}.}
  \label{fig:radar_comparison}
\end{figure}

Evaluations on \ourname~reveal a substantial gap between frontier MLLMs and true encyclopedic proficiency. As detailed in Table \ref{main result}, Gemini-3-pro leads with an F-score of 47.5\%, followed closely by \ourmodelname~(46.8\%, the top-performing open-source model). Notably, no model surpasses the 50\% threshold, underscoring the challenge of grounding the long-tail entities in our benchmark.

Category-wise analysis (see Figure \ref{fig:radar_comparison}) indicates higher proficiency in \textit{Brands} and \textit{Sports}, likely due to their over-representation in web-scale pre-training data. For instance, Gemini-3-pro achieves an F-score of 59.4 in \textit{Sports}. Conversely, \textit{Nature} and \textit{Culture} emerge as significant weaknesses. In these domains, models frequently revert to generic hypernyms (e.g., "flower" instead of specific species), which are penalized under our Granularity Alignment criteria. This suggests that while MLLMs are "pop-culture savvy," their grasp of the natural world and diverse human heritage remains shallow, necessitating more diverse data sourcing.

The discrepancy between Correct Given Attempted (CGA) and F-score serves as a probe for model honesty. GPT-5.1 exhibits a high CGA (29.3\%) but a low F-score (26.7\%), indicating a conservative strategy where the model answers only when certain. In contrast, many smaller models show low CGA, reflecting a tendency to hallucinate names for obscure entities rather than admitting ignorance. This misalignment between a model's propensity to attempt an answer and its actual accuracy suggests that current MLLMs lack a reliable internal barometer of their own knowledge boundaries, a calibration deficit we analyze in depth in Section \ref{sec:calibration}.

\subsection{Validation of Difficulty Stratification}
\label{sec: difficulty alignment}
\begin{figure}[htbp]
  \centering
  \includegraphics[width=\linewidth]{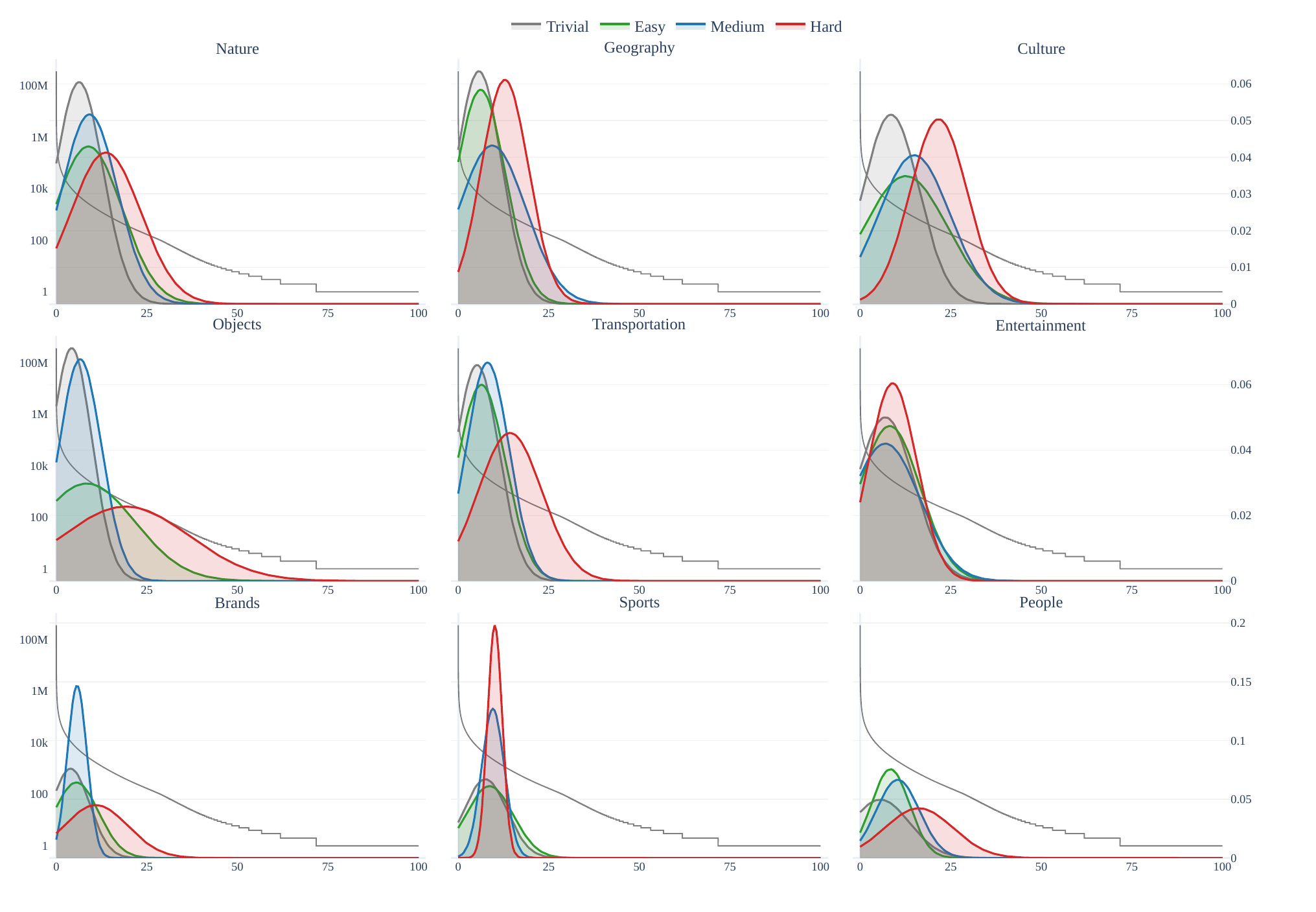}
  \caption{Entity Difficulty Distribution vs. MetaCLIP Frequency Rank Percentile across Categories. These plots illustrate the relationship between real-world entity frequency (proxied by MetaCLIP vocabulary rank percentile) and their assigned difficulty in WorldVQA. 
  The x-axis represents the percentile rank of the entity's frequency in the MetaCLIP vocabulary, where values closer to 0 indicate high-frequency (common) entities, and higher values indicate lower-frequency (rare) entities. The left y-axis corresponds to the grey line, showing the underlying exponential density distribution of MetaCLIP word frequencies, highlighting the long-tail nature of real-world knowledge. The right y-axis shows the probability density for the fitted normal distributions of the four difficulty tiers: Trivial, Easy, Medium, and Hard.
  }
  \label{fig:metaclip_result}
\end{figure}

To validate whether our Model-Performance-Based Stratification accurately reflects real-world knowledge distribution, we utilize the rank frequency of entity terms in the MetaCLIP vocabulary (\cite{xu2025demystifyingclipdata}, \cite{chuang2025metaclip2worldwide}) as a proxy for real-world prevalence. Figure \ref{fig:metaclip_result} illustrates the density distributions of entity difficulties mapped against their MetaCLIP rank percentile.

The quantitative results demonstrate a distinct positive correlation between real-world rarity and benchmark difficulty. As shown by the fitted curves, Trivial and Easy samples concentrate heavily near the zeroth percentile, indicating that current MLLMs primarily master high-frequency head entities. As difficulty escalates to Medium and Hard, the distribution peaks progressively shift rightward toward higher rank percentiles. This systematic migration confirms that the difficulty in \ourname~stems from genuine knowledge scarcity (long-tail entities) rather than confounding factors like visual ambiguity or annotation artifacts.

Furthermore, the analysis highlights \ourname's effective coverage of the knowledge spectrum. While Easy samples probe the high-frequency head, Medium and Hard categories successfully extend into critical long-tail regions often underrepresented in standard evaluations. Although minor variations exist, e.g., \textit{Brands} and \textit{People} skew slightly towards higher frequencies due to dense web coverage, the overarching trend of difficulty correlating with rarity remains robust. Thus, \ourname~provides a stratified assessment that accurately mirrors the structural complexity of real-world encyclopedic knowledge.

\subsection{Calibration Analysis}
\label{sec:calibration}

\begin{figure}[t]
\centering
\includegraphics[width=.5\linewidth]{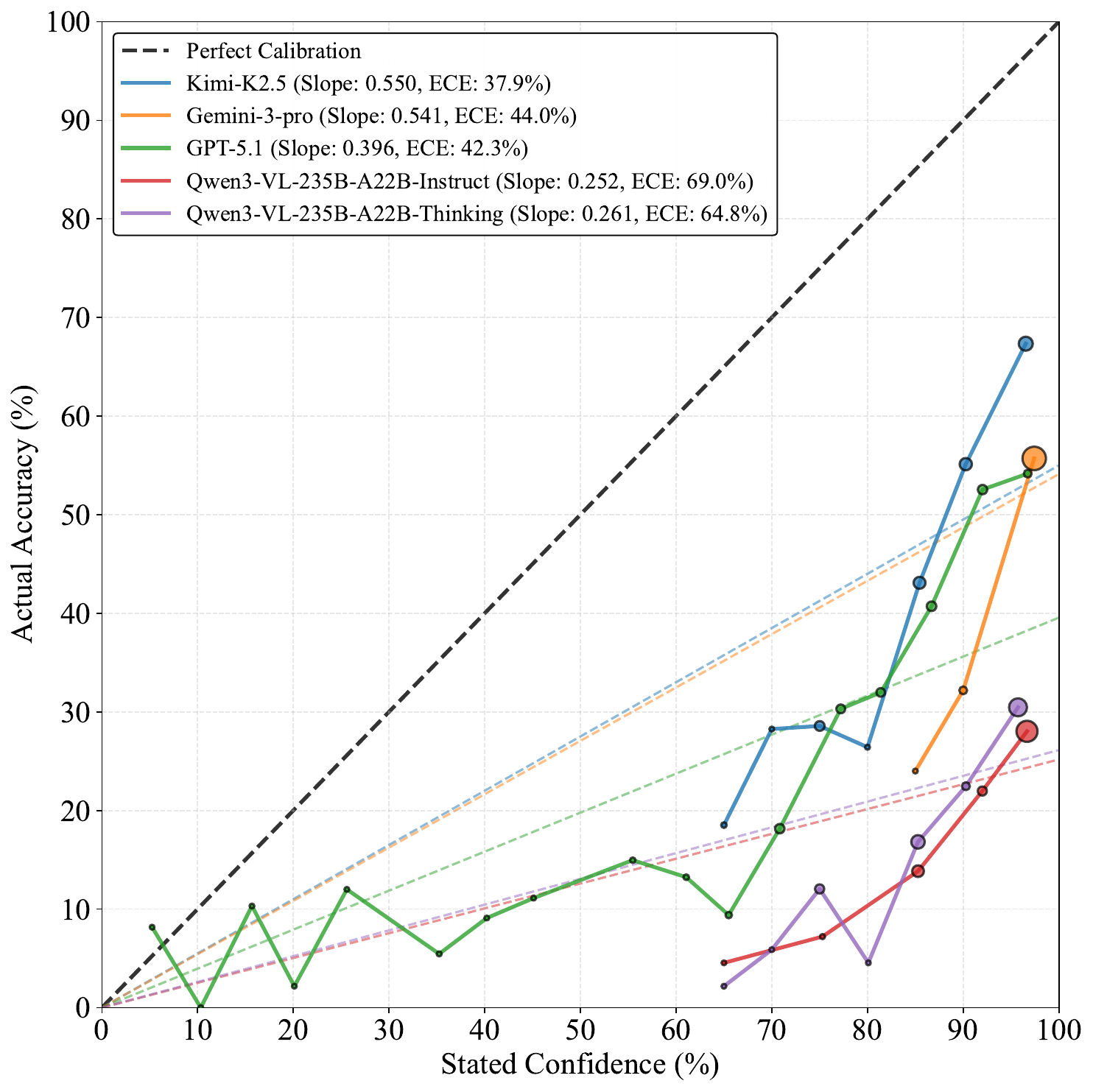}%
\hfill
\includegraphics[width=.5\linewidth]{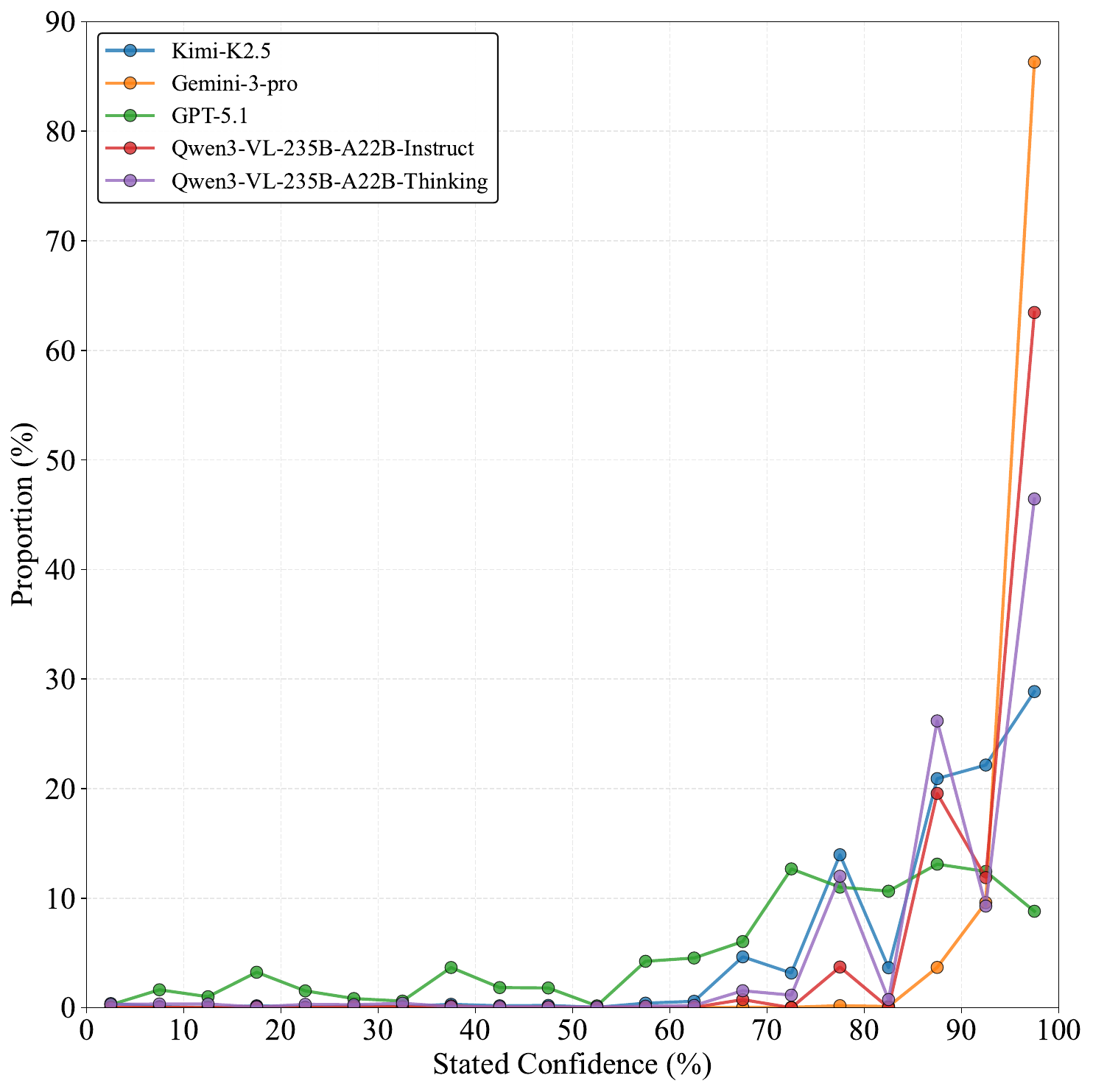}%
\caption{\textbf{Calibration and Confidence Distribution Analysis.} \textbf{Left:} Reliability diagrams plotting Actual Accuracy against Stated Confidence. To ensure statistical significance, only bins containing more than 20 samples are visualized. The size of each data point is proportional to the number of samples in that bin. The black dashed diagonal (y=x) represents perfect calibration, while colored dashed lines indicate the weighted average slope for each model. \textbf{Right:} The distribution of stated confidence scores across the full dataset (without sample thresholding). The plots reveal a severe overconfidence trend, with most models concentrating their predictions in the 90-100\% confidence range.}
\label{fig:calibration_comparison}
\end{figure}

To evaluate whether MLLMs possess a reliable sense of their own knowledge boundaries, we adopt the calibration methodology. We prompt the models to provide their best guess for each question accompanied by a confidence score on a scale of 0 to 100 (see Prompt \textit{Question with Confidence} in Appendix~\ref{sec:prompt} for prompt details). We evaluate calibration performance using two primary metrics calculated over $M$ confidence-ordered bins: 

\begin{itemize}
    \item \textbf{Expected Calibration Error (ECE)}: Measures the alignment between subjective certainty and objective accuracy (optimal ECE is 0). It is formulated as:
    $ECE = \sum_{m=1}^{M} \frac{|B_m|}{N} | \text{acc}(B_m) - \text{conf}(B_m) |$
    
    \item \textbf{Weighted Average Slope (Slope)}: Assesses the correlation between accuracy and confidence. An optimal Slope is 1.0; values significantly below 1.0 indicate systemic overconfidence.
\end{itemize}

As shown in Figure~\ref{fig:calibration_comparison}, all models exhibit severe overconfidence. \ourmodelname~achieves the best calibration (ECE: 37.9\%, Slope: 0.550), maintaining the strongest alignment between internal confidence and actual performance, outperforming both GPT and Gemini series.

Regarding confidence distribution (Figure~\ref{fig:calibration_comparison} Right), most models lack self-awareness. Gemini-3-pro shows binary behavior, assigning $\ge$ 95\% confidence in over 85\% of cases regardless of accuracy. GPT-5.1 is the only model distinguishing low confidence, offering more honest uncertainty estimates despite a slightly higher ECE. This pervasive overconfidence likely stems from a lack of uncertainty samples in training data and alignment strategies favoring assertiveness.

%% file: sec/4_Related_Work_and_Discussion.tex
\section{Related Work and Discussion}
In this paper, we propose a specialized benchmark for measuring atomic visual factuality. \ourname~sits within an extensive evaluation landscape for MLLMs. A detailed comparison between \ourname~and other representative benchmarks is provided in Table \ref{tab:benchmark_comparison}. Comprehensive suites such as MME \cite{fu2025mme}, MMBench \cite{liu2024mmbench}, SEED-Bench \cite{li2023seed}, and MMStar \cite{chen2024we} assess holistic competence, where world knowledge is often an implicit prerequisite for tasks ranging from mathematical reasoning \cite{lu2023mathvista} and text recognition \cite{liu2024ocrbench} to diagram interpretation \cite{hiippala2021ai2d} and spatial localization \cite{yu2016modeling}. While expert-level benchmarks like MMMU \cite{yue2024mmmu} and MMMU-Pro \cite{yue2025mmmu} test deep disciplinary knowledge, their emphasis on complex reasoning chains often obscures purely factual deficits. Closer to our aim are recent factuality probes like SimpleVQA \cite{cheng2025simplevqa} and VisualSimpleQA \cite{wang2025visualsimpleqa}; however, \ourname~differentiates itself by focusing on the atomic recognition of entities across a stratified taxonomy—akin to ImageNet \cite{deng2009imagenet} or LVIS \cite{gupta2019lvis}—rather than composite retrieval tasks.

We also examine reliability through the lens of calibration and hallucination, drawing on methodologies established in the language domain \cite{joshi2017triviaqa, kwiatkowski2019natural, lin2022truthfulqa, li2023halueval, wei2024measuring}. Notably, prior studies demonstrate that while pre-trained models possess latent self-knowledge \cite{kadavath2022language, lin2022teaching}, this signal is likely distorted by post-training alignment \cite{achiam2023gpt}. In the multimodal setting, evaluation has largely focused on existential or perceptual hallucination—checking for object presence \cite{rohrbach2018object, li2023evaluating} or attribute consistency \cite{guan2024hallusionbench, sun2024aligning, wang2023amber}. By isolating encyclopedic hallucination, we aim to disentangle the complex interplay between visual perception and parametric knowledge \cite{liu2024survey}, offering a granular view of model trustworthiness distinct from previous polling-based metrics.

A main limitation of \ourname~is that it measures factuality in a highly atomic setting. While this isolation allows for precise diagnosis of recognition failures, it remains an open research question whether the ability to correctly name specific entities correlates strongly with performance on complex, downstream multimodal tasks. Furthermore, we have not yet fully quantified how different Reinforcement Learning (RL) strategies specifically impact this atomic visual calibration. We hope that open-sourcing \ourname~provides the community with a rigorous baseline to investigate these dynamics and develop alignment techniques that enhance factuality without compromising uncertainty estimation.

%% file: sec/Appendix.tex
\newpage
\section*{Appendix}
\section{Prompts}
\label{sec:prompt}
\begin{tcolorbox}[
title=Question Prompt, 
colback=white, 
colframe=black!70, 
arc=5pt,
fonttitle=\bfseries\ttfamily]

Please provide as much detail as possible in your answer. \{question\}

\end{tcolorbox}

\begin{tcolorbox}[
title=Question with Confidence Prompt, 
colback=white, 
colframe=black!70, 
arc=5pt,
fonttitle=\bfseries\ttfamily]

Please provide as much detail as possible in your answer and give your best guess. \{question\}

At the end, please provide a confidence score (0-100\%).

\end{tcolorbox}

\begin{tcolorbox}[
title=Visual Audit Prompt, 
colback=white, 
colframe=black!70, 
arc=5pt,
fonttitle=\bfseries\ttfamily]
You are a strict "Visual Fact-Checker." Your task is to determine whether the provided original image can serve as the sole, conclusive evidence to support the given answer to a question. Your core principle is: "Believe only what is seen; reject all speculation." You must judge if the image provides unique, conclusive, and exclusive evidence to validate the answer.

The input format is:

"""

Question: \{question\}

Ground Truth Answer: \{ground\_truth\_answer\}

"""

Evaluation Steps and Guidelines:

1. Analyze Question and Answer: Carefully read the Question (Q) and Answer (A) to understand the core knowledge or facts involved.

2. Verify Visual Evidence: Closely examine the image. Search for direct visual evidence that confirms the claims made in Answer (A).

\begin{itemize}[label=--, leftmargin=2em]
    \item Point 1 - Clarity: Is the information in the image clear enough to identify Answer (A)?
    \item Point 2 - Uniqueness/Exclusivity: Does the visual information support \textit{only} Answer (A) for Question (Q), or could it support other reasonable alternatives?
    \item Point 3 - Completeness: Does the image contain all the necessary information required to answer Question (Q)?
\end{itemize}
   
3. \*\*Write Reasoning (Mandatory):\*\* Detail how you arrived at your conclusion based on specific image details. Clearly list the key visual evidence that supports or refutes the answer.

\*\*Final Conclusion Format Requirements:\*\*

judge\_reason: [Your detailed reasoning]

judge\_result: [A, B, or C]

Based on your reasoning, you must select exactly one of the following three options as your judge\_result:

A. Determinable (The image fully supports the answer)

B. Inconclusive (Cannot be clearly determined; requires more detail)

C. Incorrect (The image contradicts the answer)
\end{tcolorbox}

\begin{tcolorbox}[
title=Judge Prompt, 
colback=white, 
colframe=black!70, 
arc=5pt,
fonttitle=\bfseries\ttfamily]

\textbf{Role}

You are an expert judge specialized in evaluating the correctness of answers. Your task is to assess whether a model-generated answer is correct based on a given question, the model's response, and the ground truth answer.

\textbf{Task: Evaluate Answer Correctness}

Please classify the model's response into one of the following three categories. Ignore differences in formatting, punctuation, language (Chinese vs. English), or abbreviations/full names. Focus strictly on the \*\*core semantics\*\* and the \*\*level of detail (granularity)\*\*:

\begin{enumerate}[label=\arabic*., leftmargin=2em]
    \item \textbf{Correct}:
    \begin{itemize}[label=--, leftmargin=2em]
        \item The model answer contains the core information of the ground truth.
        \item The model answer is semantically consistent with the ground truth and contains no contradictions.
        \item The granularity of the model answer is equal to or finer than the ground truth.
        \item Extra irrelevant information is allowed as long as it does not conflict with the ground truth.
    \end{itemize}

    \item \textbf{Incorrect}:
    \begin{itemize}[label=--, leftmargin=2em]
        \item The model answer provides information that contradicts the ground truth.
        \item The model answer provides the wrong specific entity, value, or description.
        \item The granularity of the model answer is coarser than the ground truth, leading to incomplete or insufficiently specific information.
        \item Even if the model expresses uncertainty but follows up with a wrong answer (e.g., "I'm not sure, maybe it's B" when the truth is A), it is considered Incorrect.
    \end{itemize}

    \item \textbf{Unattempted}:
    \begin{itemize}[label=--, leftmargin=2em]
        \item The model explicitly states it does not know the answer (e.g., "I don't know," "I cannot answer this question").
        \item The model suggests the user search elsewhere (e.g., "Please search the internet").
        \item The model answer contains no information from the ground truth but provides no incorrect or contradictory information.
    \end{itemize}
\end{enumerate}

\textbf{Output Format}

Please strictly follow this two-line format for your output:
\begin{enumerate}[label=\arabic*., leftmargin=2em]
\item \*\*Evaluation\*\*: [A brief explanation of your reasoning]

\item \*\*Label\*\*: [Final classification: "Correct", "Incorrect", or "Unattempted"]
\end{enumerate}

\textbf{Examples}

Input:

\*\*Example 1 (Correct - Finer Granularity)\*\*

Input:

"""

Question: What weather phenomenon is in the image?

Model Answer: Based on the visual evidence in the image, the weather phenomenon shown is a \*\*severe storm with extremely high winds\*\*, most likely a \*\*tornado\*\* or a very powerful \*\*hurricane/typhoon\*\*.

Ground Truth Answer: High winds

"""

Evaluation: The ground truth is "high winds," and a "tornado" is a more specific and granular type of high wind. The semantics are correct and the detail is finer.

Label: Correct

... (cases of incorrect and unattempted)

\textbf{Current Task}

Input:

"""

Question: \{question\}

Model Answer: \{model\_answer\}

Ground Truth Answer: \{ground\_truth\_answer\}

"""

Evaluation:
\end{tcolorbox}

\clearpage 
\section{WorldVQA Showcases}
\label{sec:showcases}
\subsection{Nature \& Environment} %
\begin{center}
    \nopagebreak 
    \includegraphics[width=\linewidth, trim=0 510 0 0, clip]{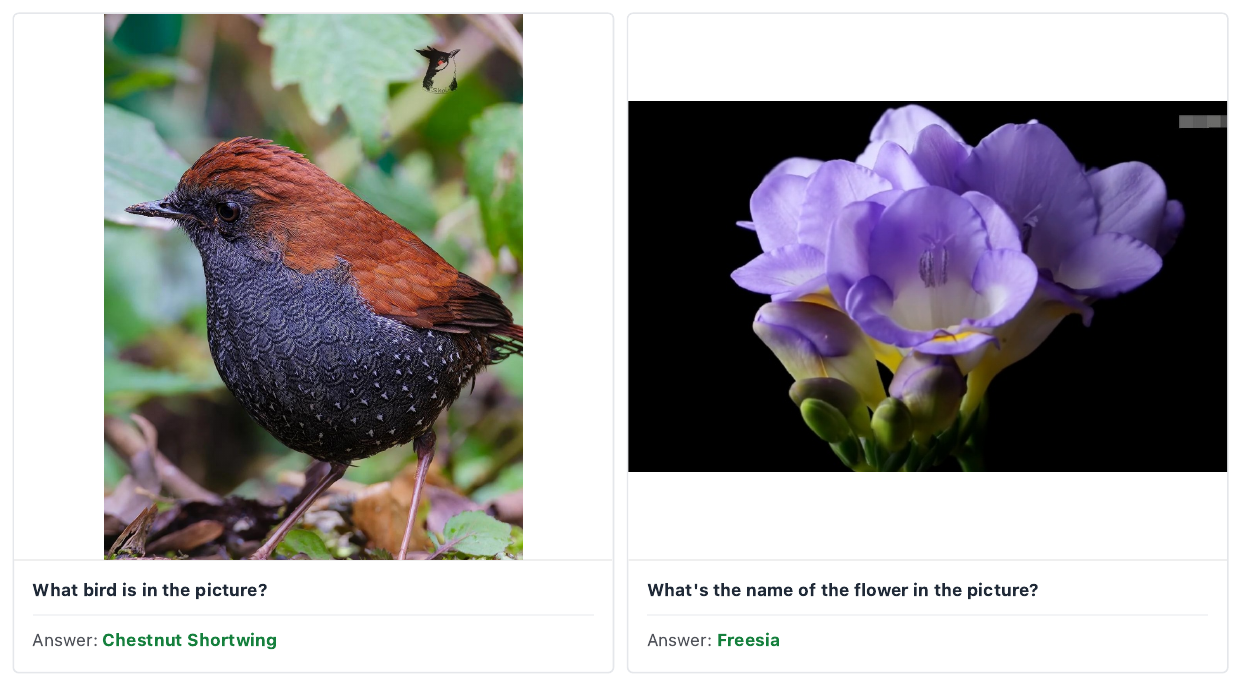}
\end{center}

\subsection{Locations \& Architecture} %
\begin{center}
    \nopagebreak
    \includegraphics[width=\linewidth, trim=0 510 0 0, clip]{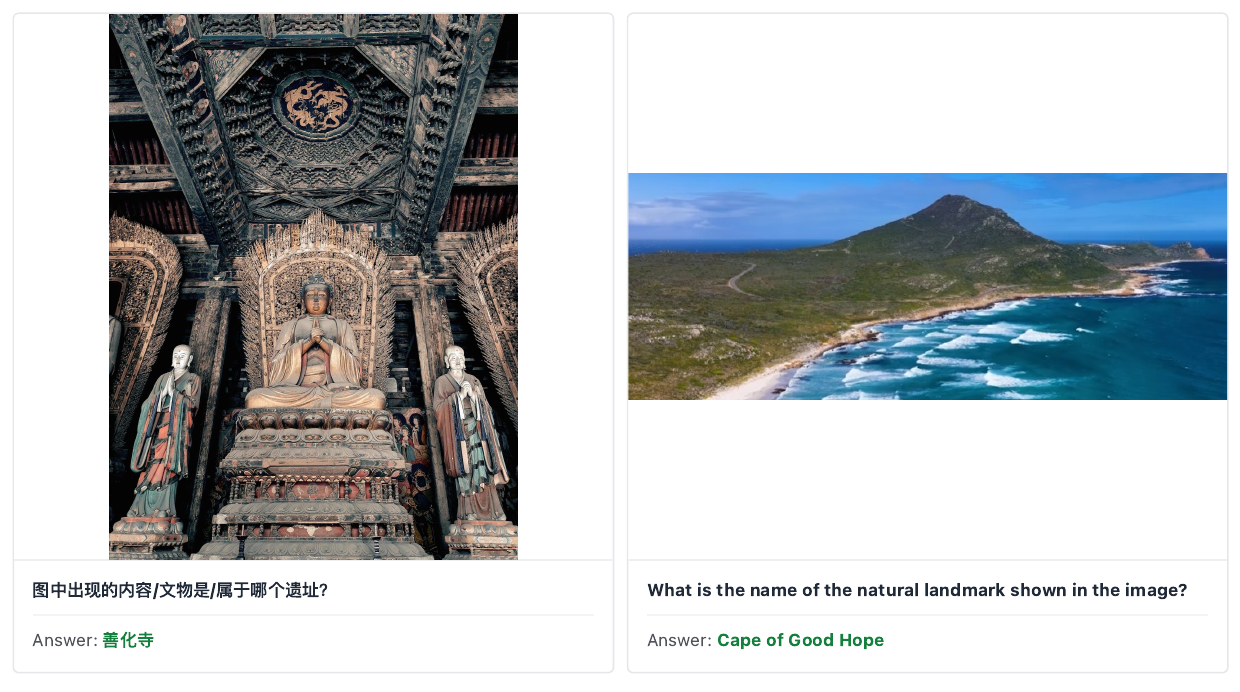}
\end{center}

\subsection{Culture, Arts \& Crafts} %
\begin{center}
    \nopagebreak
    \includegraphics[width=\linewidth, trim=0 510 0 0, clip]{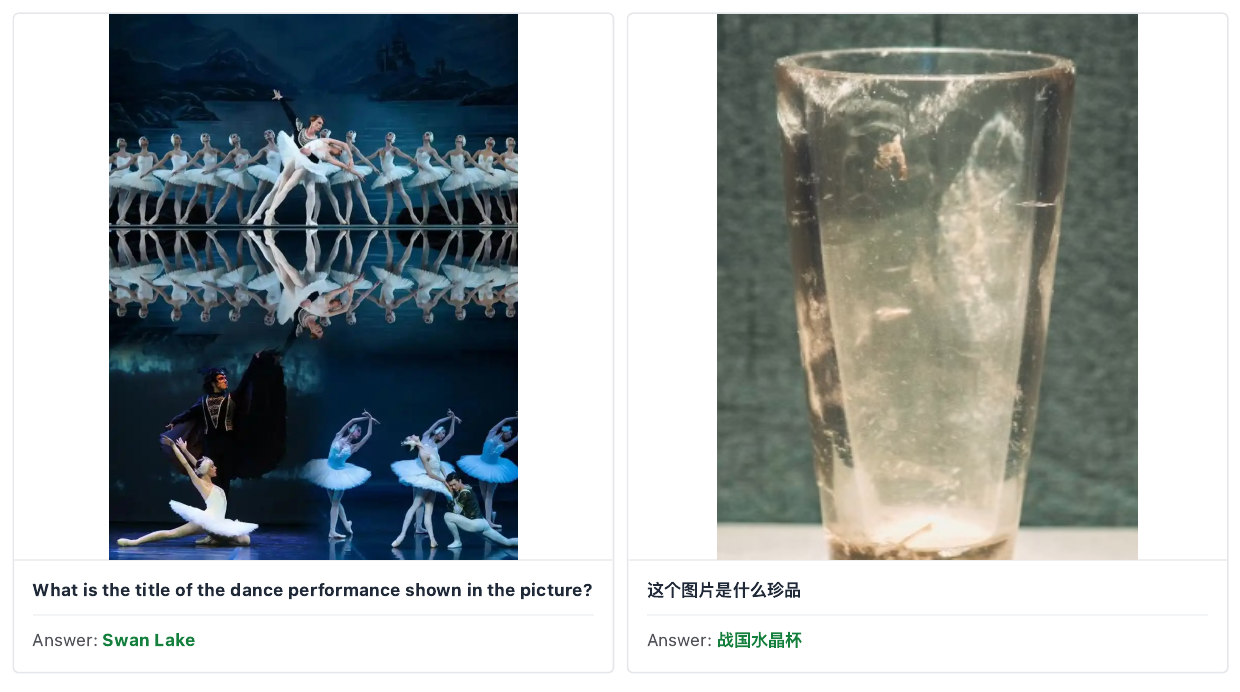}
\end{center}

\subsection{Objects \& Products} %
\begin{center}
    \nopagebreak
    \includegraphics[width=\linewidth, trim=0 510 0 0, clip]{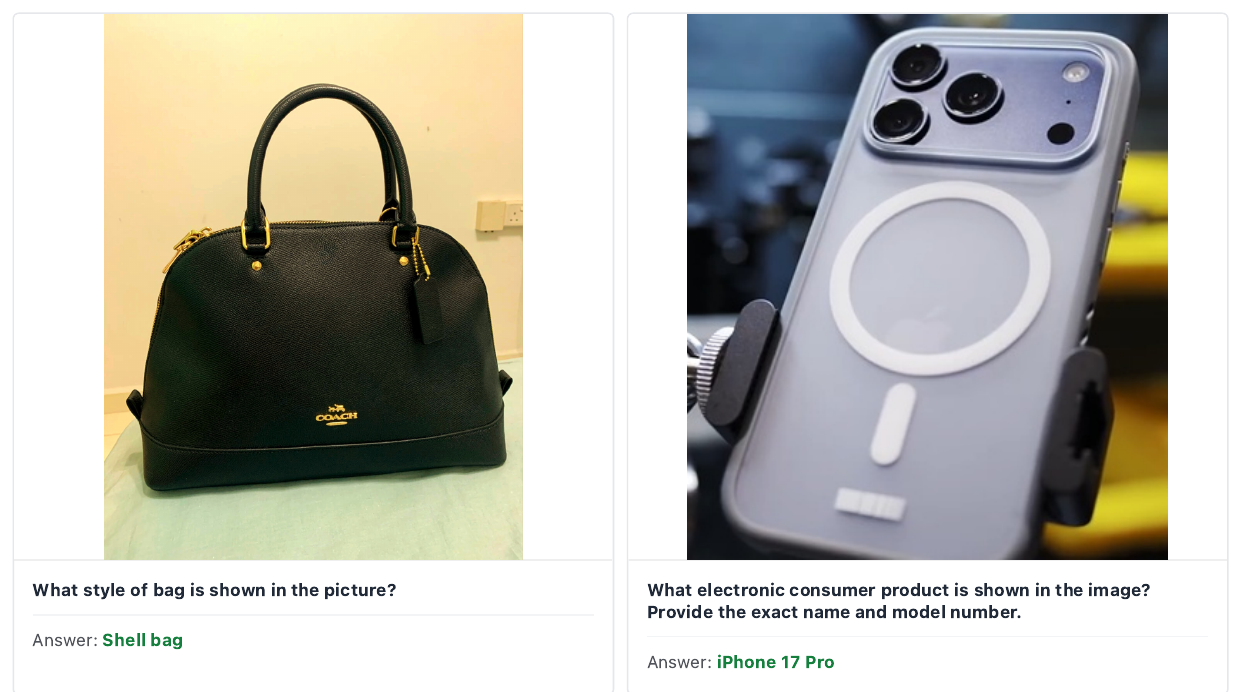}
\end{center}

\subsection{Vehicles, Craft \& Transportation} %
\begin{center}
    \nopagebreak
    \includegraphics[width=\linewidth, trim=0 510 0 0, clip]{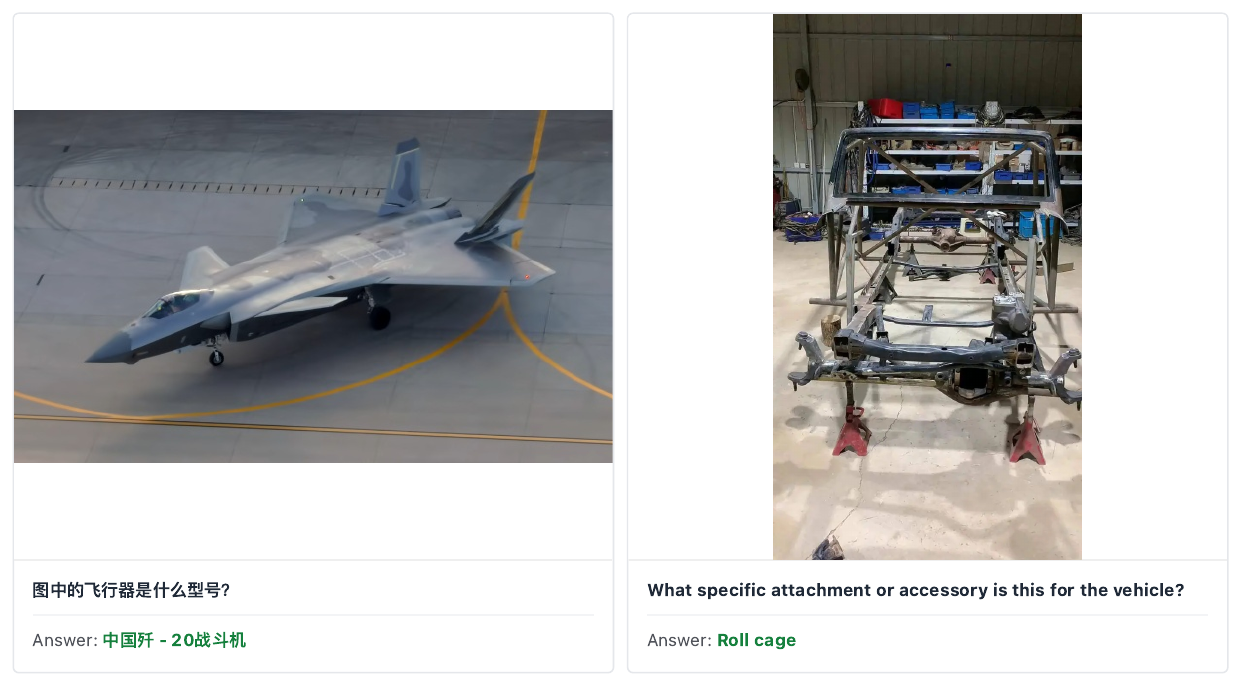}
\end{center}

\subsection{Entertainment, Media \& Gaming} %
\begin{center}
    \nopagebreak
    \includegraphics[width=\linewidth, trim=0 510 0 0, clip]{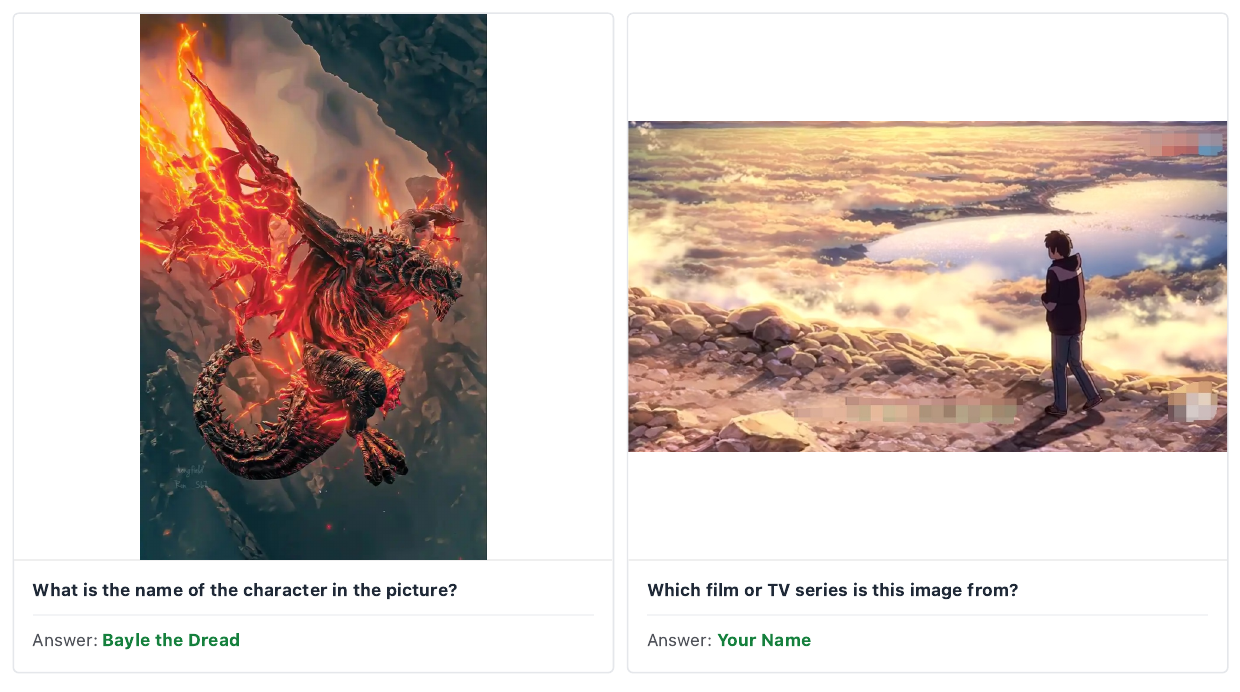}
\end{center}

\subsection{Brands, Logos \& Graphic Design} %
\begin{center}
    \nopagebreak
    \includegraphics[width=\linewidth, trim=0 510 0 0, clip]{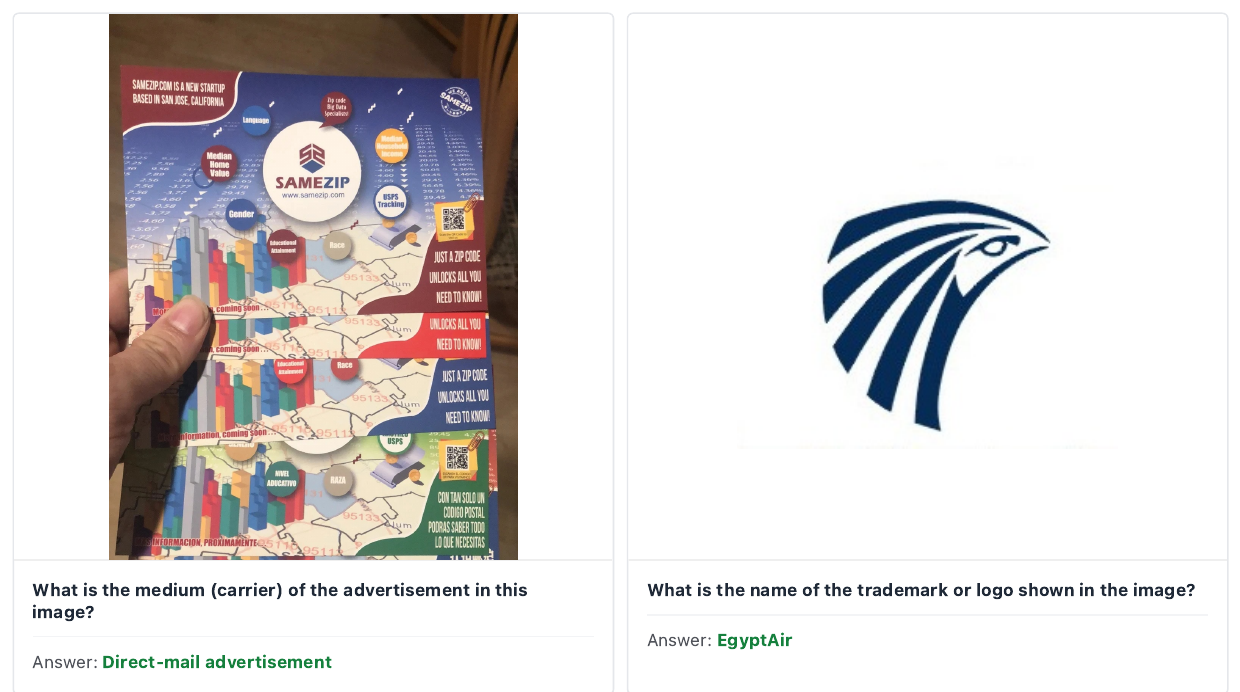}
\end{center}

\subsection{Sports, Gear \& Venues} %
\begin{center}
    \nopagebreak
    \includegraphics[width=\linewidth, trim=0 510 0 0, clip]{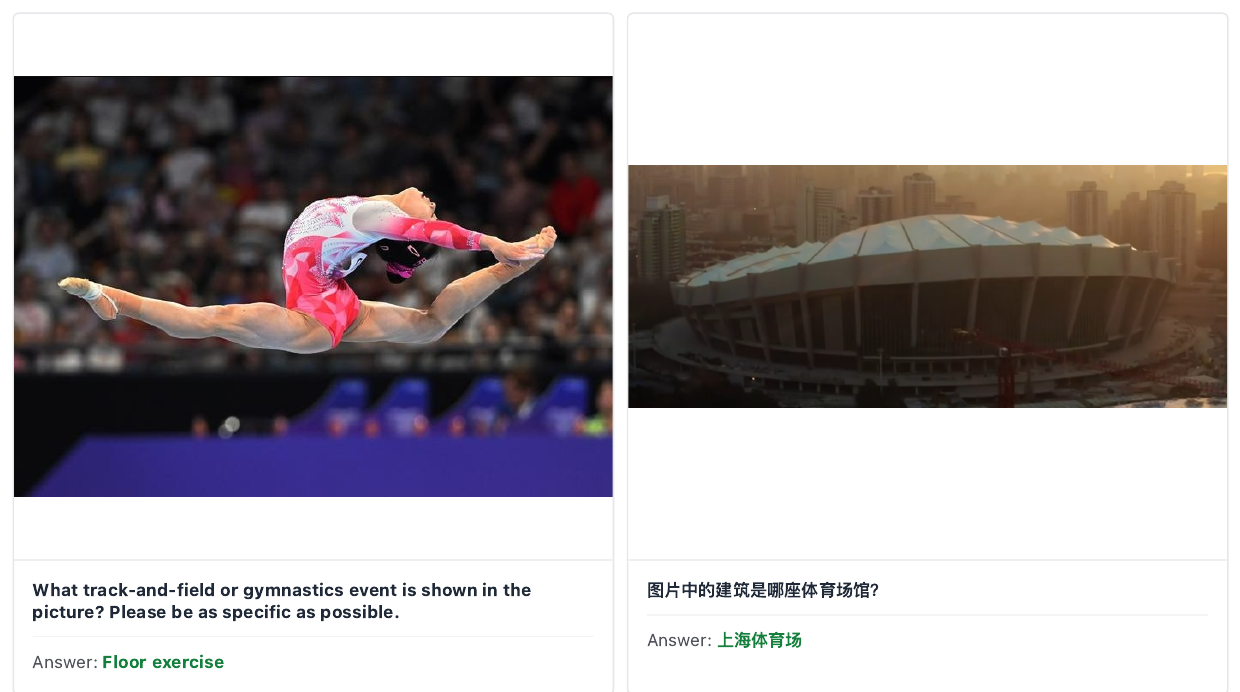}
\end{center}